# Evaluating the application of NLP tools in mainstream participatory budgeting processes in Scotland

Jonathan Davies

Computer Science, Warwick University, Coventry, UK, jonathan.davies.1@warwick.ac.uk

Miguel Arana-Catania

Computer Science, Warwick University, Coventry, UK, miguel.arana-catania@warwick.ac.uk

Rob Procter

Computer Science, Warwick University, Coventry, UK & Alan Turing Institute for Data Science and AI, rob.procter@warwick.ac.uk

Felix-Anselm van Lier

Government Outcomes Lab, Oxford University, UK, felix-anselm.vanlier@bsg.ox.ac.uk

Yulan He

Computer Science, Warwick University, Coventry, UK, yulan.he@warwick.ac.uk

In recent years participatory budgeting (PB) in Scotland has grown from a handful of community-led processes to a movement supported by local and national government. This is epitomized by an agreement between the Scottish Government and the Convention of Scottish Local Authorities (COSLA) that at least 1% of local authority budgets will be subject to PB. This ongoing research paper explores the challenges that emerge from this 'scaling up' or 'mainstreaming' across the 32 local authorities that make up Scotland. The main objective is to evaluate local authority use of the digital platform Consul, which applies Natural Language Processing (NLP) to address these challenges. This project adopts a qualitative longitudinal design with interviews, observations of PB processes, and analysis of the digital platform data. Thematic analysis is employed to capture the major issues and themes which emerge. Longitudinal analysis then explores how these evolve over time. The potential for 32 live study sites provides a unique opportunity to explore discrete political and social contexts which materialize and allow for a deeper dive into the challenges and issues that may exist, something a wider cross-sectional study would miss. Initial results show that issues and challenges which come from scaling up may be tackled using NLP technology which, in a previous controlled use case-based evaluation, has shown to improve the effectiveness of citizen participation.

## 1 Introduction

As disenchantment with political institutions grows many governments are turning to innovative practices that attempt to deepen citizen participation in decision-making processes [5, 19]. One such practice is participatory budgeting (PB) which strives to empower local communities and tackle democratic deficit by involving citizens in budget allocation [18]. As a growing global movement, PB is experiencing an additional innovative turn - the use of digital platforms to further enhance democratic processes and the effectiveness of government [18, 2].

In Scotland, PB processes have proliferated from a handful of community-led processes to an exercise embedded within the mainstream of Scottish policy and politics. This is exemplified by an agreement between the Convention of Scottish Local Authorities (COSLA) and Scottish Government that 1% of council funds (at least £100 million) will be subject to PB in 2021. Further, a roadmap is in place for progressively rolling out PB for over £10bn of public sector spending in Scotland. This ongoing research paper explores the complexities that come from this scaling up and the use of a digital participation platform and NLP technology to tackle such problems.

Following the work of Harkins et al. [12], Escobar et al. [9] and O'Hagan et al. [17] we begin with a review of the existing state of PB in Scotland. We then explore the application of digital PB, the use of natural language processing (NLP) technology applied to digital participation and introduce the intended research methodology. Initial results suggest that the inevitable issues of capacity and information overload which come from scaling up may be tackled using NLP technology which, in a previous controlled, use case-based evaluation, has shown to improve the effectiveness of citizen participation and collective intelligence [2]. This may then allow space for greater deliberation, meaningful co-production and a transformation of power, factors important for effective PB.

## 2   Participatory budgeting

PB was first developed in Porto Alegre, Brazil in 1989 as a budgetary decision-making process. Presented as a new way of doing politics between the Partido dos Trabalhadores (the Workers' Party) and its citizens [4] it was implemented to tackle democratic deficits and better support marginalised communities [1]. Since its inception PB has evolved from a local process to a global movement undergoing many adaptations along the way. The Participatory Budgeting World Atlas estimates there now to be between 11,690-11,825 PB cases worldwide [7]. This proliferation complicates any attempt to apply a single definition to all processes. For consistency, this paper follows Cabannes [3] and uses the definition coined by PB's creators from Porto Alegre:

> "It is a mechanism (or a process) by which the population defines the destination of part or the totality of public resources. Participatory budgeting is a process of direct democracy, universal and voluntary, through which the population can discuss and define the public budget and policies. PB combines direct democracy and representative democracy" (Uribatam de Souza in [3] p. 257)

By disassembling this definition, we can bring attention to the characteristics which are present across all PB initiatives and recognise its distinction from other innovative practices:

- The population – PB engages directly with the population as citizens, placing them at the centre of the process. Here we see a separation of PB from other participatory processes which are designed to engage with certain expert groups, panels, or a cross-section of the public. Smith [19] notes that experts or stakeholders are also members of the public. However, PB is innovative in the fact it engages with citizens because they are citizens. Escobar [8] expands the notion of citizen, transcending the common or legalistic definition to include typically excluded groups such as refugees or children.
- Defining a part or the totality of public resources – Participatory processes are often designed to collate ideas with action dependent on top-level agreement. PB instead involves a commitment to either the entire, or a specific portion, of an institution's budget.
- Through a process combining direct and representative democracy – Key actors of the representative system play a central role in the PB process, adapting to direct democratic challenges, managing the process, and making use of the results. Through this lens PB becomes something more than simply a reallocation of budgets but, as Escobar [8] illustrates, a form of co-production. This is a collective process of shared power, creating new connections between citizens, political representatives, and local government officials.

## 3   Participatory budgeting in Scotland

The emergence of PB in Scotland can be attributed to various political, institutional, and social factors. Firstly, issues of national identity and dissatisfaction with democratic structures have arisen as a result of political events such as the Scottish independence referendum in 2014, the EU 'Brexit' referendum in 2016 and the recent UK General Elections [12]. These tensions are exacerbated by pre-existing institutional factors. The 32 local authorities which make up Scotland serve a population of 5.4 million, resulting in an average of 168,750 citizens per local authority (compared with the EU average of 5,615 citizens) and a ratio of 1 elected councillor each representing 4,270 Scottish citizens (compared with a ratio of 1:700 in Spain, 1:500 in Germany, and 1:400 in Finland) [9]. This results in Scotland having "the largest average population per basic unit of local government of any developed country" [9 p. 312]. Finally, despite potential for disengagement between citizens and local authorities there is a strong social desire for greater participation in local decision making (see [14] and [13] for two recent surveys on participation) and civic engagement [11]. This desire for participation complemented with the political and institutional factors has generated a turn towards public service reform, community empowerment, and desire for democratic renewal which PB has been at the heart of [9].

Early PB processes in Scotland were often supported by the Scottish Government or local authorities but can be largely characterised by community grant-making, emerging organically from grass-root organisations [9]. This first wave is commonly referred to as 'first generation PB' [12]. The grassroots momentum soon gained increasing political support from the Scottish Government following the Community Choices Fund, a national development programme to underpin the introduction of PB in partnership with local authorities [17]. This juncture represents a move to what is commonly described as 'second generation PB' or mainstreaming, the embedding of PB across Scotland. This was further strengthened in 2017 when the Scottish Government and COSLA agreed that at least 1% of local authority budget will be decided by citizens and communities through PB. With this juncture we can return to the characteristics explored in the earlier definition. Firstly, the mainstreaming of PB moves beyond community grant making and places all citizens, or the population, at the centre of the process. Secondly, local authorities have committed to a set amount of budget to be used. Third, the embedding of the process is not just the reallocation of funds but instead creates a space for co-production through new connections between citizens and local authority officials.

## 4   Digital participatory budgeting

In order to involve as much of the population as possible in the PB process, governments around the world are implementing digital platforms for participation. This has led to the development of important projects such as the free software Consul project. This collaborative project between governments around the world has led to use cases such as that of the city of Madrid, one of the largest in the world with 100 million euros decided by the citizens every year. The project was inspired by the Icelandic platform "My Neighbourhood", the Icelandic portal using the free software technology of Better Reykjavík, which was the first major global success story of this type of platform, launched as a consequence of the Icelandic crisis of 2008-2011 [5]. The wide popularisation of the Consul project has led Porto Alegre to implement it locally, thus receiving the digital legacy of the participatory project originated in that city.

The Consul platform has been chosen in Scotland to carry out the digital transition of participatory budgeting. Although in this work we are studying the comparative use between local authorities in Scotland, it is very informative to have as a reference its use by other governments in other countries, such as the previously mentioned case of Madrid, or other smaller cities such as Groningen in the Netherlands.

Unlike other participatory processes, PB processes typically allow for the collection of large numbers of openly drafted proposals which have to be evaluated by both government and citizens. This means that the complexity of the process grows exponentially with respect to the number of participants in the process. This is not the case in offline participation processes where the participation divide means that the number of participants is small. Although most of the digital processes implemented worldwide are still small-scale processes, this opens the door to problems of information overload as they develop. Some specific cases are already beginning to show these characteristics, for example in the city of Madrid where more than 90,000 participants registered in 2018.

To tackle the problem of information overload in participatory budgeting from its earliest appearances NLP technologies have recently started to be applied to participatory budgeting processes. Our latest study [2] shows how such technology can be applied in these digital participation processes, making them more effective and facilitating participation and the capacity for information analysis and interaction during the process.

In the previous case study these NLP technologies were analysed in a controlled use case. On this occasion these same technologies will be applied to live cases in participatory processes in Scotland providing a unique opportunity to analyse these effects.

# 5 NLP applied to digital participation

The NLP technology to be used in the Scottish case study is the same that was applied in the controlled study mentioned above. It has been designed to enhance aspects of the participation process and interaction with the platform where the amount of information available makes it difficult for users to process, or where users find it difficult to interact with each other. This is critical for information and deliberation to flow properly and to generate collective intelligence amongst users. However, as we have pointed out, the complexity of these tasks grows rapidly in proportion to the number of participants in the processes.

To facilitate these processes, the following NLP techniques were implemented on the platform [2]:

- Topic modelling on the proposals. This made it possible to identify the global categories of the proposals, assign tags according to them to each proposal, and define a distance between proposals that allowed them to be grouped together. The latter made it possible to list and link in that space all the similar existing proposals when accessing one of them. The general categorisation of proposals solves in a simple and coherent way the problem of the wide taxonomic variety generated by the users themselves, and also makes it possible to generate a global map of the topics dealt with throughout all the proposals.
- Text summarisation on the comments. Extractive automatic summarisation techniques made it possible to summarise all the comments associated with each proposal and display them next to the proposal. This facilitates interaction as users can have a general idea of the opinions around each proposal, allowing them to easily receive synthetic information about it, as well as to decide which debates to explore in more detail.
- Topic modelling on the user generated content. This technique allows the use of publicly generated content by users to identify which users have similar combinations of interests and invite them to collaborate together. This facilitates interaction between users, which is the main basis for the deliberative processes that take place in the platform.

These developments therefore facilitate more effective and intelligent interaction both between users and with the content created in the deliberative process. Additionally, they can also be used by public decision-makers in charge of the participatory processes to perform a global analysis of the opinions and ideas of the participants. It is common in participatory platforms to develop data analysis technology that is only accessible to administrators in a top-down design. However, in this case the results of the analysis are always public and open, and are designed in a horizontal way, focused on facilitating participation and interaction between users. The goal is not to extract intelligence from users, but to make the processes themselves smarter.

# 6 Research methodology

The use of the Consul platform and NLP technology by up to 32 local authorities provides a unique opportunity. Each council, with distinct demographics, administrative structures, and political landscape, brings discrete reasons as to why they wish to employ digital PB, how they will use the platform, and the spaces that need to be shaped to ensure a cultural change. To therefore understand the effectiveness of the platform on PB in Scotland, we must explore the social and political contexts which individually emerge from each authority before and during its use.

This project adopts a qualitative longitudinal design using a cohort study. This involves observations of a particular group (in this case, local authorities using Consul), repeated over a period of time. A longitudinal study is often recognised as a key research method for measuring change and forming causal interpretations [15]. It moves beyond a static analysis by introducing a temporal aspect to study which can document changes in attitudes that emerge, something not possible with traditional cross-sectional data.

In-depth secondary research will first be carried out to provide contextual information on the current state of PB in Scotland. We will then observe local authorities through the attendance of meetings related to PB and semi-structured interviews with key actors. With the potential for 32 live study sites, it is essential that any methods of data recording and classification are identical. As such, a hybrid process of inductive and deductive thematic analysis [10] will be employed to identify and report on any common themes which exist or may emerge from our observations over time. This approach allows for the following: (1) A process of deductive coding, based on initial secondary research, will provide a good baseline to determine PB's current state across the 32 local authorities in Scotland before the implementation of the Consul platform; (2) an inductive approach then allows for emergence of new themes as and when they occur. These new themes indicate the influence and impact of the Consul platform on the PB process through the lens of local authorities. As there is no strict time frame for when each authority will adopt the platform, this engagement and analysis will unfold organically and will continuously occur before and during the implementation of the Consul platform.

# 7 Initial results

Secondary research on the evaluations of 'first generation PB' have highlighted a number of important challenges to consider when moving towards a second generation, or mainstream, model.

In a review of first generation PB, Harkins et al. [12] recognise that PB is at its strongest when processes are bespoke and tailored, adapting to community contexts, priorities, and aspirations. They suggest that meaningful dialogue and deliberation between actors should feature more prominently in the design and implementation of PB processes and become a key component in the evaluation of the democratic quality of PB. The use of digital engagement platforms to aid the PB processes was also recommended, something which is supported by the Democratic Society in their review of digital tools and PB [6].

O'Hagan et al. [17] echo Harkins et al. [12] recommendations. In an evaluation report of PB activity in Scotland between 2016-2018, they suggest early iterations have been dominated by either transactional models - a time limited exchange of financial resources and transference models - an exchange of decision-making power from state to individual. They conclude that, although these two models have important benefits, a transformative model (one of co-production and empowerment) is superior for mainstream PB. This requires "a significant improvement in the deliberative opportunities and processes for supporting participation in decision making at local level and at the level of council budgets" [17 p. 10]. In an interim report of PB in Scotland, O'Hagan et al. [16] also suggest that activities to date have represented a significant resource commitment by local authorities with existing staff taking on additional workloads.

Initial interviews with local authority officials have confirmed the concerns raised from secondary research. Emerging themes include a strain on the local authority workforce, high resource cost, a cultural change resulting in resistance from internal staff, and a lack of deliberation or discussion during the process itself.

# 8 Potential impact

Done well, PB can provide open and transparent engagement, promote social capital, and empower citizens. Secondary research has shown there is a strong desire for the mainstreaming of PB in Scotland and appetite for greater participation in decision making. However, a growth in participation inevitably leads to an increase of citizen input which may overwhelm both institutions and citizens [2]. This is particularly true for Scotland whose local authorities are far from 'local' with just 32 councils serving a population of 5.4 million [9], a concern raised in our initial interviews. This may negatively affect the PB process which could fuel negative sentiments around local government, generate mistrust (from both citizens and institutions of representative democracy) and further isolate already marginalised groups. To ensure an effective and enduring process, PB must therefore be designed to tackle potential information overload, allow for greater deliberation, meaningful co-production, and a transformation of power without placing greater strain on the local authority workforce [9, 17].

A previous analysis of the Consul platform, which used NLP in a laboratory-based study, showed an improvement in the participatory process through affording participants improved information analysis and interaction [2]. Therefore, the application of this technology for mainstream PB processes in Scotland is a recommended step to tackle the issues raised above and provide space for greater deliberation, meaningful co-production, and a transformation of power. This will allow COSLA to provide a better platform to its member councils and allow these councils to provide a better service to their citizens and communities. If the pilot is successful, the enhanced Consul platform will become the vehicle for further roll out of PB processes in Scotland on decisions on over £10bn of public sector spending by 2022. The Consul platform with NLP technology will also be made available for Consul's existing user base (135 institutions in 35 countries) and future users to download and install. The next stage of this research will be to continue observations and semi-structured interviews with local authorities across Scotland to compare and analyse the effects of the Consul platform within councils.

## ACKNOWLEDGMENTS

This research has been supported by the Alan Turing Institute for Data Science and AI (grant no. EP/N510129/1) and by the EPSRC Impact Acceleration Fund (grant no. G.CSAA.0705).